\pdfoutput=1

\documentclass[11pt]{article}

\usepackage[]{emnlp2021}

\usepackage{microtype}
\usepackage{makecell}
\usepackage{multirow}
\usepackage{amsmath} 
\usepackage{adjustbox}
\usepackage{xpatch}  
\usepackage{arydshln} 
\newcommand\numberthis{\addtocounter{equation}{1}\tag{\theequation}}

\xpretocmd{\footnote}{\unskip}{}{}
\usepackage{times}
\usepackage{latexsym}

\usepackage[T1]{fontenc}

\usepackage[utf8]{inputenc}

\usepackage{microtype}

\title{Scheduled Sampling Based on Decoding Steps for \\ Neural Machine Translation}

\author{
  Yijin Liu\textsuperscript{12},
  Fandong Meng\textsuperscript{2}, 
  Yufeng Chen\textsuperscript{1} 
  Jinan Xu\textsuperscript{1}
  and Jie Zhou\textsuperscript{2} \\
  \textsuperscript{1}Beijing Jiaotong University, China \\
  \textsuperscript{2}Pattern Recognition Center, WeChat AI, Tencent Inc, China \\
  \texttt{\{yijinliu, fandongmeng, withtomzhou\}@tencent.com} \\
  \texttt{\{jaxu,chenyf\}@bjtu.edu.cn} \\
}

\begin{document}
\maketitle

\begin{abstract}
Scheduled sampling is widely used to mitigate the exposure bias problem for neural machine translation. Its core motivation is to simulate the inference scene during training by replacing ground-truth tokens with predicted tokens, thus bridging the gap between training and inference. However, vanilla scheduled sampling is merely based on training steps and equally treats all decoding steps. Namely, it simulates an inference scene with uniform error rates, which disobeys the real inference scene, where larger decoding steps usually have higher error rates due to error accumulations. To alleviate the above discrepancy, we propose scheduled sampling methods based on decoding steps, increasing the selection chance of predicted tokens with the growth of decoding steps. Consequently, we can more realistically simulate the inference scene during training, thus better bridging the gap between training and inference. Moreover, we investigate scheduled sampling based on both training steps and decoding steps for further improvements. Experimentally, our approaches significantly outperform the Transformer baseline and vanilla scheduled sampling on three large-scale WMT tasks. Additionally, 
our approaches also generalize well to the text summarization task on two popular benchmarks.
\end{abstract}

\begin{figure}[t!]
\begin{center}
     \scalebox{0.42}{
      \includegraphics[width=1\textwidth]{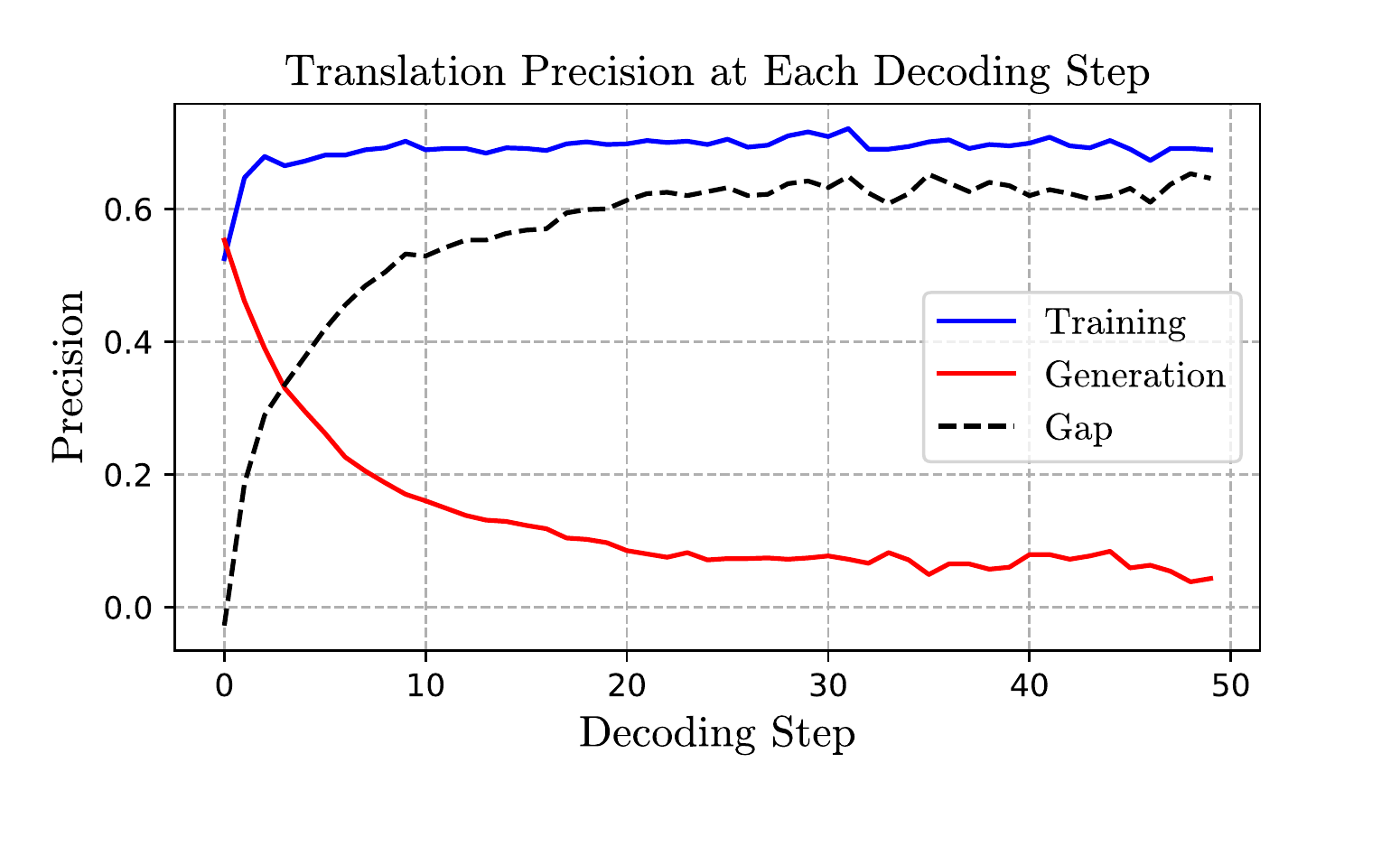}
      } 
   \caption[caption tag]{
   The translation precision for training (blue line) and inference (red line) at each decoding step.
  The gap between training and inference (black line) increases rapidly with the growth of decoding steps. We randomly sample 100k training data from WMT 2014 EN-DE and report the average precision of 1k tokens for each decoding step\protect\footnotemark.
   }
  \label{fig:training_inference_prec_gap}  
  \vspace{-8pt}
 \end{center} 
\end{figure}

\footnotetext{
To calculate the precision for training, we strictly match predicted tokens with ground-truth tokens word by word. 
When inference, we relax the strict matching to the fuzzy matching within a local window of size 3, and truncate or pad hypotheses to the same length of golden references. 
We also explore $n$-gram matching in preliminary experiments and observe analogical results with different $n$. For simplicity, we use the above unigram matching to calculate the translation precision (similarly for the error rate) in all experiments.}

\section{Introduction}
\label{sec:introduction}
Neural Machine Translation (NMT) has made promising progress in recent years \cite{seq2seq_2014,bahdanau_nmt_2015,transformer_2017}. Generally, NMT models are trained to maximize the likelihood of next token given previous golden tokens as inputs, {\em i.e.,} teacher forcing \cite{teacher_forcing_2016}. However, at the inference stage, golden tokens are unavailable. The model is exposed to an unseen data distribution generated by itself. This discrepancy between training and inference is named as the \textit{exposure bias} problem \cite{mixer_2015}. With the growth of decoding steps, such discrepancy becomes more problematic due to error accumulations  \cite{sdnmt_2019,bi_direction_2020} (shown in Figure \ref{fig:training_inference_prec_gap}).

Many techniques have been proposed to alleviate the exposure bias problem. To our knowledge, they mainly fall into two categories. The one is sentence-level training, which treats the sentence-level metric ({\em e.g.,} BLEU) as a reward, and directly maximizes the expected rewards of generated sequences \cite{mixer_2015,mrt_2016,rl_related_2017,pang_rl_2021}. Although intuitive, they generally suffer from slow and unstable training due to the high variance of policy gradients and the credit assignment problem \cite{credit_assignment_1985,optimize_beam_2016,rl_related1_2018,rl_related2_2018}. Another category is sampling-based approaches, aiming to simulate the data distribution of the inference scene during training. Scheduled sampling \cite{ss_2015} is a representative method, which samples tokens between golden references and model predictions with a scheduled probability. 
\citet{zhang_bridging_2019} further refine the sampling candidates by beam search.
\citet{ss_transformer_2019} and \citet{parallel_ss_2019} extend scheduled sampling to the Transformer with a novel two-pass decoder architecture. \citet{confidence_ss_2021} develop a more fine-grained sampling strategy according to the model confidence.

Although these sampling-based approaches have been shown effective and training efficient, there still exists an essential issue in their sampling strategies. In the real inference scene, the nature of sequential predictions quickly accumulates errors along with decoding steps, which yields higher error rates for larger decoding steps \cite{sdnmt_2019,bi_direction_2020} (Figure \ref{fig:training_inference_prec_gap}). However, most sampling-based approaches are merely based on training steps and equally treat all decoding steps\footnote{For clarity in this paper, `training steps' refer to the number of parameter updates and `decoding steps' refer to the index of decoded tokens on the decoder side.}. Namely, they simulate an inference scene with uniform error rates along with  decoding steps, which is inconsistent with the real inference scene.

To alleviate this inconsistent issue, we propose scheduled sampling methods based on decoding steps, which increases the selection chance of predicted tokens with the growth of decoding steps. In this way, we can more realistically simulate the inference scene during training, thus better bridging the gap between training and inference. Furthermore, we investigate scheduled sampling based on both training steps and decoding steps, which yields further improvements. It indicates that our proposals are complementary with existing studies.
Additionally, we provide in-depth analyses on the necessity of our proposals from the perspective of translation error rates and accumulated errors. Experimentally, our approaches significantly outperform the Transformer baseline by 1.08, 1.08, and 1.27 BLEU points on WMT 2014 English-German, WMT 2014 English-French, and WMT 2019 Chinese-English, respectively. 
When comparing with the stronger vanilla scheduled sampling method, 
our approaches bring further improvements by 0.58, 0.62, and 0.55 BLEU points on these WMT tasks, respectively. Moreover, our approaches generalize well to the text summarization task and achieve consistently better performance on two popular benchmarks, {\em i.e.,} CNN/DailyMail \cite{cnndw_dataset_2017} and Gigaword \cite{ggw_dataset_2015}. 

The main contributions of this paper can be summarized as follows\footnote{Codes are available at \url{https://github.com/Adaxry/ss_on_decoding_steps.}}:
\begin{itemize}
    \item To the best of our knowledge, we are the first that propose scheduled sampling methods based on decoding steps from the perspective of simulating the distribution of real translation errors, and provide in-depth analyses on the necessity of our proposals.
    \item We investigate scheduled sampling based on both training steps and decoding steps, which yields further improvements, suggesting that our proposals complement existing studies.
    \item Experiments on three large-scale WMT tasks and two popular text summarization tasks confirm the effectiveness and generalizability of our approaches. 
    \item Analyses indicate our approaches can better simulate the inference scene during training and significantly outperform existing studies.
    
\end{itemize}

\section{Background}
\subsection{Neural Machine Translation}
Given a pair of source language $\mathbf{X} = \{x_1, x_2, \cdots, x_m \}$ and target language $\mathbf{Y} = \{y_1, y_2, \cdots, y_n \}$, neural machine translation aims to model the following translation probability:
\begin{align*}
     P(\mathbf{Y}|\mathbf{X}) 
     & = \sum_{t=1}^{n}{\log P(y_{t} | \mathbf{y}_{<t}, \mathbf{X}, \theta)}
     \numberthis
     \label{equ:nmt_define}
\end{align*}    
where $t$ is the index of target tokens, $\mathbf{y}_{<t}$ is the partial translation before $y_t$, and $\theta$ is model parameter. In the training stage, $\mathbf{y}_{<t}$ are ground-truth tokens, and this procedure is also known as teacher forcing. The translation model is generally trained with maximum likelihood estimation (MLE).

\begin{figure}[t!]
\begin{center}
     \scalebox{0.475}{
      \includegraphics[width=1\textwidth]{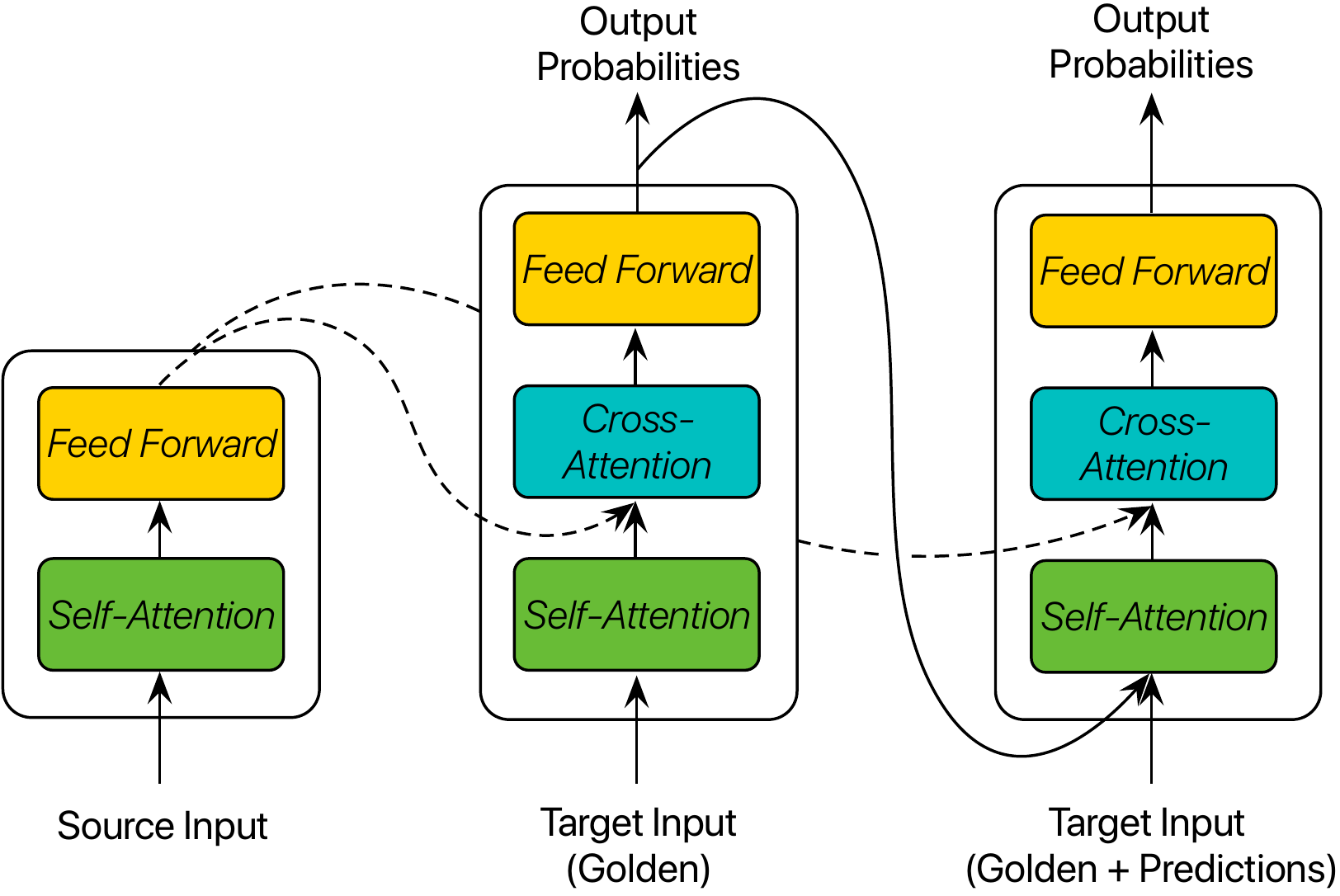}
      } 
      \caption{
      Scheduled sampling for the transformer with a two-pass decoder at training.
      } 
      \label{fig:ss_for_transformer}  
      \vspace{-8pt}
 \end{center} 
\end{figure}
\subsection{Scheduled Sampling for the Transformer}
 Scheduled sampling is initially designed for Recurrent Neural Networks \cite{ss_2015}, and further modifications are needed when applied to the Transformer \cite{ss_transformer_2019,parallel_ss_2019}.
As shown in Figure \ref{fig:ss_for_transformer}, we follow the two-pass decoder architecture for the training of Transformers. 
In the first pass, the model conducts the same as a standard NMT model. Its predictions are used to simulate the inference scene \footnote{Following \citeauthor{prediction_softmax_2017} (\citeyear{prediction_softmax_2017}), model predictions are the weighted sum of target embeddings over output probabilities. As model predictions cause a mismatch with golden tokens, they can simulate translation errors of the inference scene.}.
In the second pass,  the decoder's inputs $\widetilde{\mathbf{y}}_{<t}$ are sampled from predictions of the first pass and ground-truth tokens with a certain probability. 
Finally, predictions of the second pass are used to calculate the cross-entropy loss, and Equation (\ref{equ:nmt_define}) is modified as follow:
\begin{align*}
     P(\mathbf{Y}|\mathbf{X}) = \sum_{t=1}^{n}{log P(y_{t} | \widetilde{\mathbf{y}}_{<t}, \mathbf{X}, \theta)}
     \numberthis
     \label{equ:ss_for_transformer}
\end{align*}    
Note that the two decoders are identical and share the same parameters during training. At inference, only the first decoder is used, that is just the standard Transformer. 
How to schedule the above probability of sampling tokens for training is the key point, which is  we aim to improve in this paper. 

\subsection{Decay Strategies Based on Training Steps}
\label{sec:decay_strategies_on_training_steps}
\begin{figure}[t!]
\begin{center}
     \scalebox{0.46}{
      \includegraphics[width=1\textwidth]{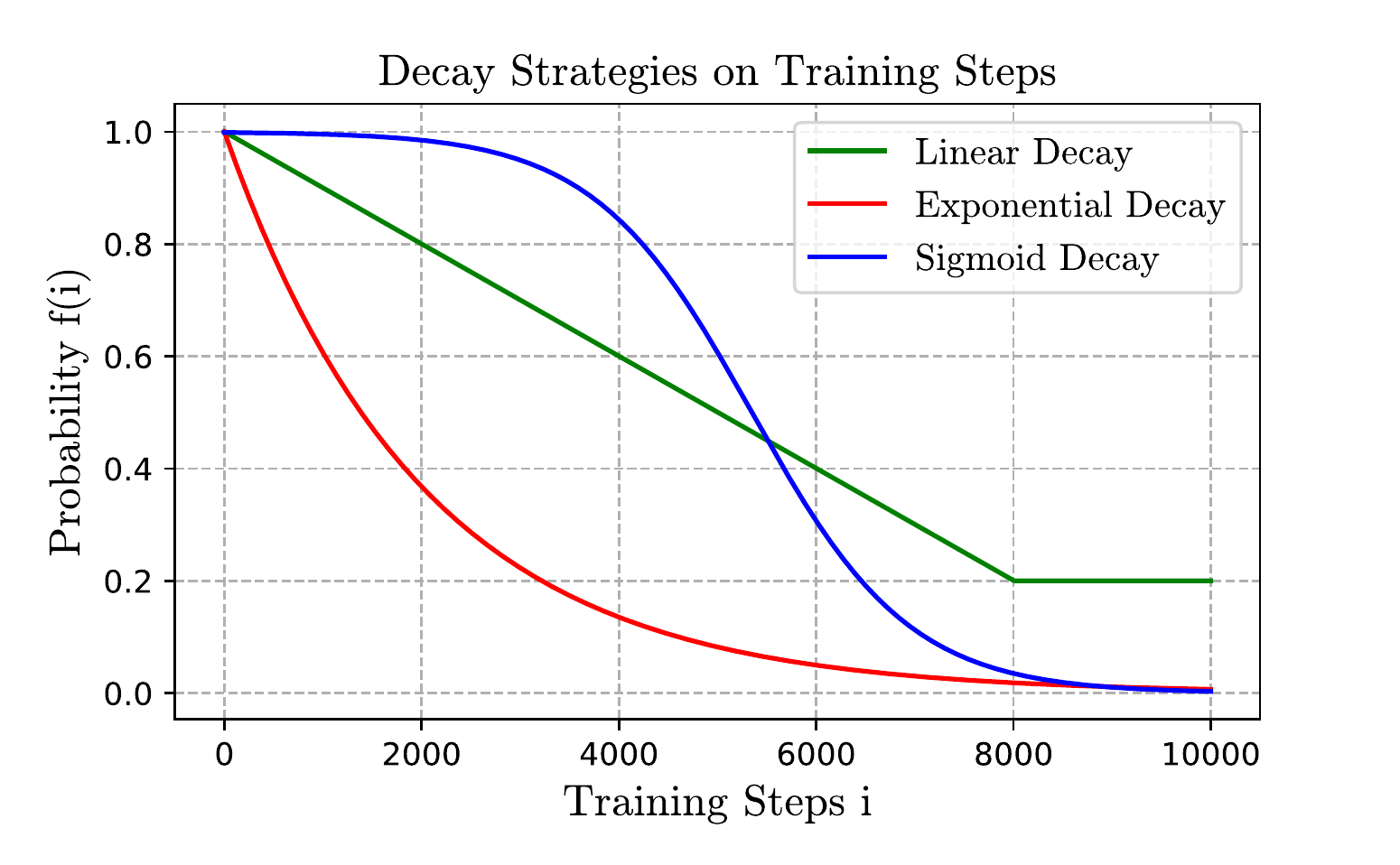}
      } 
      \caption{
      Examples of different decay strategies $f(i)$.
      } 
      \vspace{-8pt}
      \label{fig:training_step_decay_strategy}  
 \end{center} 
\end{figure}

Existing schedule strategies are based on training steps \cite{ss_2015,zhang_bridging_2019}.
At the $i$-th training step, the probability of sampling golden tokens $f(i)$ is calculated as follow:
\begin{itemize}
\item Linear Decay: $f(i) = \max(\epsilon , ki + b)$, where $\epsilon$ is the minimum value, and $k < 0$ and $b$ is respectively the slope and offset of the decay.
\item Exponential Decay: $f(i) = k^i$, where $k < 1$ is the radix to adjust the decay.
\item Sigmoid Decay\footnote{For simplicity, we abbreviate the `Inverse Sigmoid decay' \cite{ss_2015} to `Sigmoid decay.'}: $f(i) = \frac{k}{k + e^{\frac{i}{k}}}$, where $e$ is the mathematical constant, and $k \geq 1$ is a hyperparameter to adjust the decay. 
\end{itemize}
We draw some examples for different decay strategies based on training steps in Figure \ref{fig:training_step_decay_strategy}.

\section{Approaches}
\subsection{Definitions and the Overview}
At the training stage, in the input of the second-pass decoder, each token is sampled either from the golden token or the predicted token by the first-pass decoder.
For clarity, we only define the probability of sampling golden tokens, {\em e.g.,} $f(i)$, and use $1-f(i)$ to represent the probability of sampling predicted tokens.
Specifically, we define the probability of sampling golden tokens as $f(i)$ when sampling based on the training step $i$, as $g(t)$ when sampling based on the decoding step $t$, and as $h(i, t)$ when sampling based on both training steps and decoding steps.
In this paper, when we mention a scheduled strategy, it is about the probability of sampling golden tokens at the model training stage. 
In this section, we firstly point out the drawback of merely sampling based on training steps. Secondly, we describe how to appropriately sample based on decoding steps. Finally, we explore whether sampling based on both training steps and decoding steps  can complement each other.

\subsection{Sampling Based on Training Steps}
As the number of the training step $i$ increases, the model should be exposed to its own predictions more frequently. Thus a decay strategy for sampling golden tokens $f(i)$ (in Section \ref{sec:decay_strategies_on_training_steps}) is generally used in existing studies \cite{ss_2015,zhang_bridging_2019}.
At a specific training step $i$, given a target sentence, $f(i)$ is only related to $i$ and equally conducts the same sampling probability for all decoding steps.
Therefore, $f(i)$ simulates an inference scene with uniform error rates and still remains a gap with the real inference scene.

\subsection{Sampling Based on Decoding Steps}
We take a further step to bridge the above gap $f(i)$ left.
Specifically, we propose sampling based on decoding steps and schedule the sampling probability $g(t)$ under the guidance of real translation errors. 
As mentioned earlier (Figure \ref{fig:training_inference_prec_gap}), translation error rates are growing rapidly along with decoding steps in the real inference stage.
To more realistically simulate such error distributions of the real inference scene during training, we expose more model predictions for larger decoding steps and more golden tokens for smaller decoding steps.
Thus it is intuitive to apply a decay strategy for sampling golden tokens based on the number of decoding steps $t$. Specifically, we directly inherit above decay strategies (Section \ref{sec:decay_strategies_on_training_steps}) for training steps $f(i)$ to $g(t)$ with a different set of hyperparameters (listed in Table \ref{table:hyperparameters_for_ss}). 

\begin{figure}[t!]
\begin{center}
     \scalebox{0.465}{
      \includegraphics[width=1\textwidth]{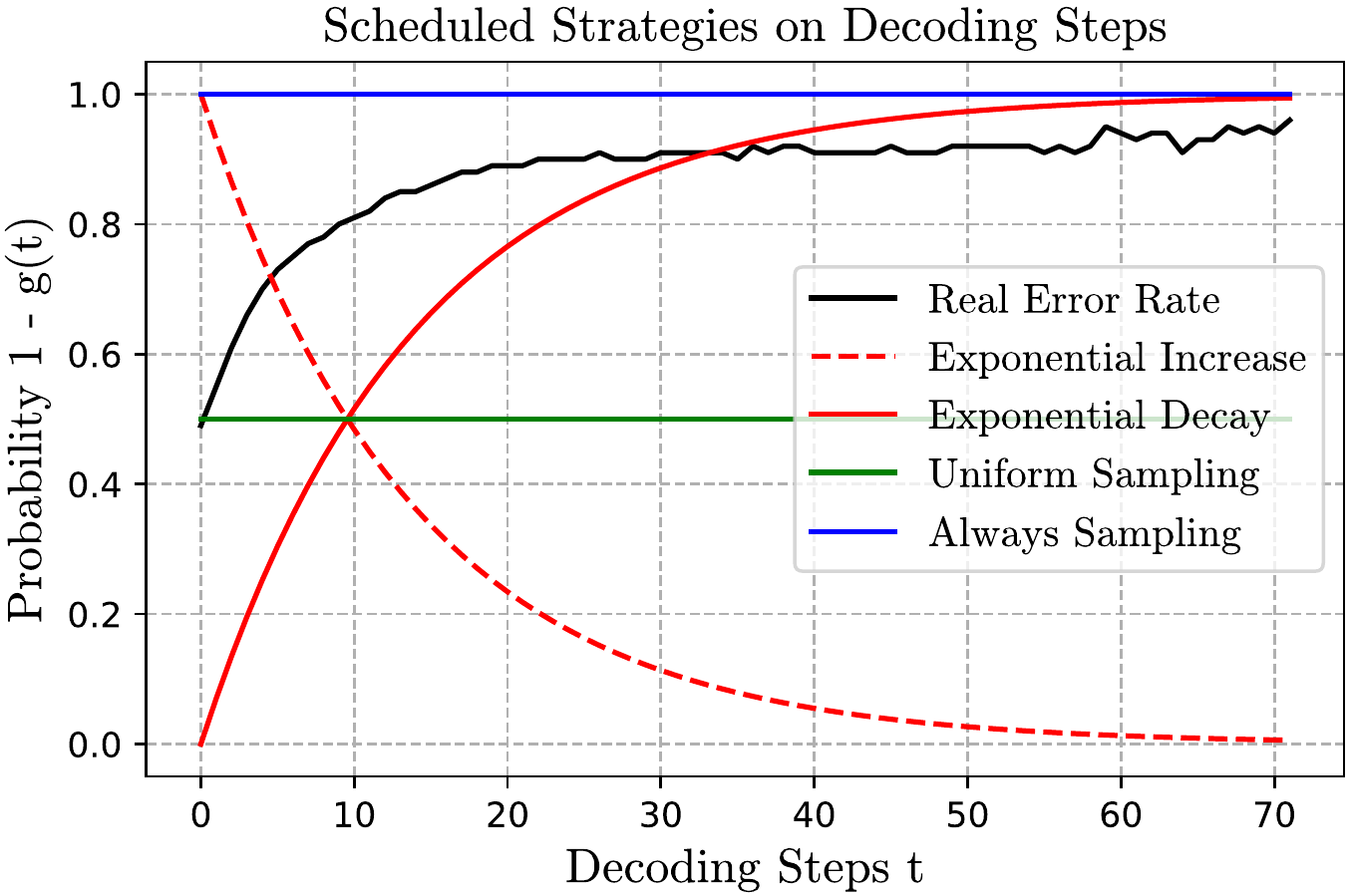}
      } 
      \caption{
      Examples of different strategies for $1-g(t)$ based on the decoding step $t$. The black line refers to the real error rate calculated by unigram matching.
      } 
      \vspace{-2pt}
      \label{fig:decoding_step_decay_strategy}  
 \end{center} 
\end{figure}

To rigorously validate the necessity and effectiveness of our proposals, 
we further conduct the following method variants for comparisons:
\begin{itemize}
\item Always Sampling: This model always samples from its own predictions.
\item Uniform Sampling: This model randomly samples golden tokens with a uniform probability (0.5 in our experiments).
\item Increase Strategies: These models reverse decay strategies to increase strategies, {\em i.e.,} $g(t) \rightarrow 1-g(t)$.
\end{itemize}
We draw some representative strategies \footnote{For brevity, we omit linear and sigmoid strategies, which show analogical trends with the exponential strategy.} in Figure \ref{fig:decoding_step_decay_strategy}. 
Both `Always Sampling' (blue line) and `Uniform Sampling' (green line) parallel to the x-axis, namely irrelevant with $t$. They serve as baseline models to verify whether a scheduled strategy is necessary on the dimension of $t$.
The exponential decay (solid red line) shows a similar trend with the real error rate (black line): the larger decoding steps and the higher error rates. 
On the other hand, the exponential increase (dashed red line) is entirely contrary to the real error rate. 
However, we cannot take it for granted that the exponential increase is inappropriate, as it can still simulate the error accumulation phenomenon\footnote{We will elaborately analyze effects of different schedule strategies in Section \ref{sec:effect_of_schedule_strategies}.}.
Therefore, merely comparing error rates is not enough. We need to step deeper into the dimension of error accumulations for further comparisons.

\paragraph{Error Accumulations.}
\begin{figure}[t!]
\begin{center}
     \scalebox{0.45}{
      \includegraphics[width=1\textwidth]{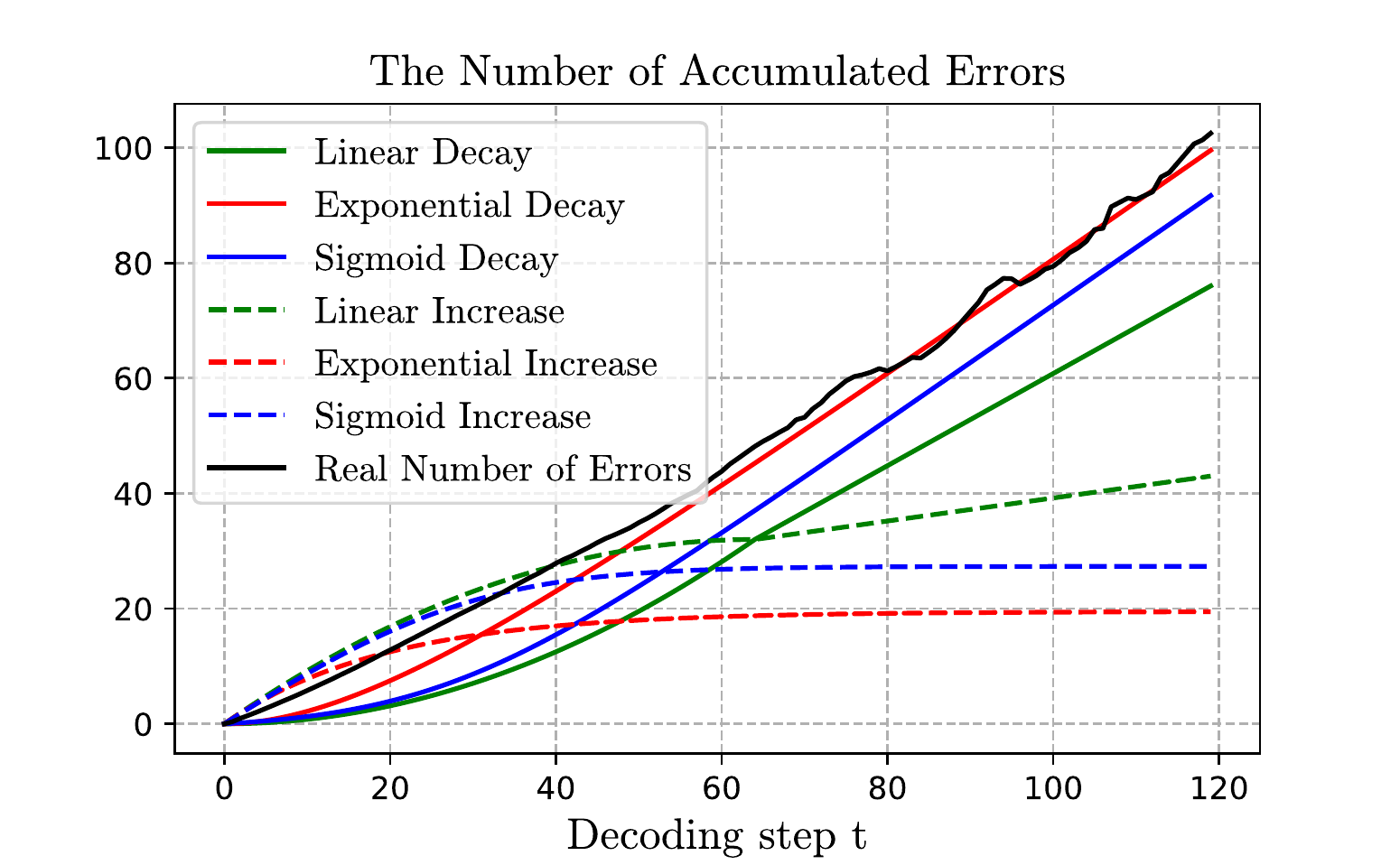}
      } 
      \vspace{-2pt}
      \caption{
      Simulated accumulated errors for different strategies. Solid lines refer to decay strategies, dashed lines refer to increase strategies, and the black line represents the real number of accumulated errors calculated by unigram matching.
      } 
      \vspace{-2pt}
      \label{fig:accumulated_errors}
 \end{center} 
\end{figure}

\begin{figure*}[t!]
\begin{center}
     \scalebox{0.99}{
      \includegraphics[width=1\textwidth]{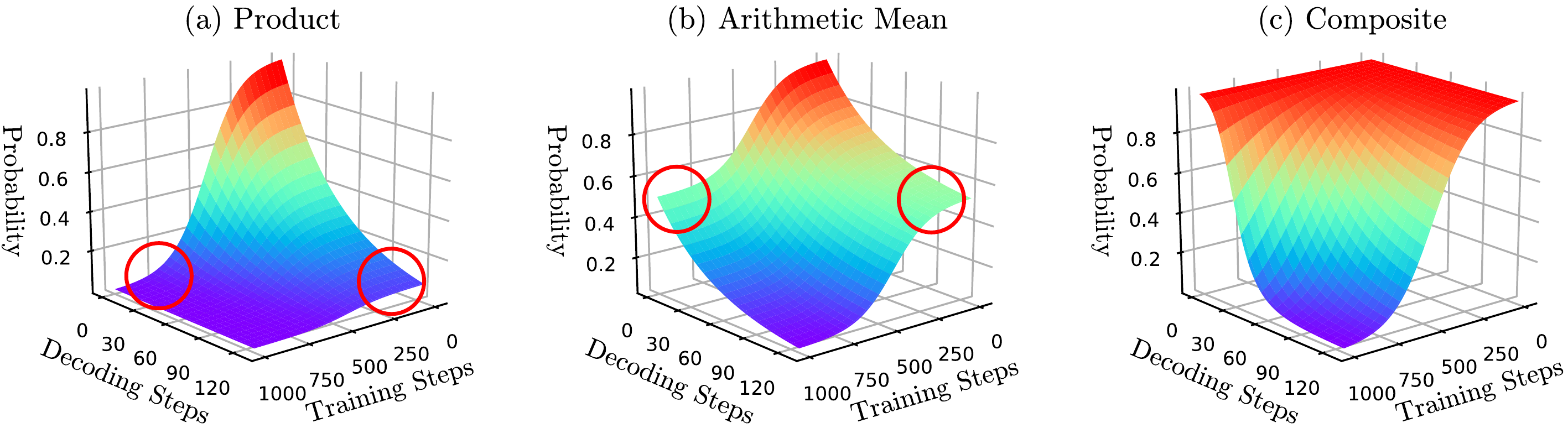}
      } 
      \caption{
      Examples for different $h(i, t)$. The wavelengths of colors represent the probability of sampling golden tokens. 
      Namely, the closer the color to the red, the greater the probability.
      Red circles are for the sake of highlights.
      } 
 \label{fig:ss_on_both}  
 \vspace{-3pt}
 \end{center} 
\end{figure*}

At the decoding step $t$, the number of accumulated errors $accum(t)$ is the definite integral of the probability of sampling model predictions $1-g(t)$:
\begin{align*}
     accum(t) = \int_{0}^{t} (1-g(x)) \,dx
     \numberthis
     \label{equ:integration}
\end{align*}    
As shown in Figure \ref{fig:accumulated_errors}, $accum(t)$ is a monotonically increasing function, which can simulate the error accumulation phenomenon no matter which kind of scheduled strategy $g(t)$ during training. 
Nevertheless, we observe that different strategies show different speeds and distributions for simulating error accumulations. For instance, decay strategies (solid lines) show a slower speed at the beginning of decoding steps and then rapidly accumulate errors with the growth of decoding steps, which is analogous with the real inference scene (black line).
However, increase strategies (dashed lines) are just on the contrary. They simulate a distribution with lots of errors at the beginning and an almost fixed number of errors in following decoding steps.
Moreover, although different decay strategies show similar trends for simulating  error accumulations in the training stage, the degrees of their approximations with real error numbers are still different.
We will further validate whether the proximity is closely related to the final performance in Section \ref{sec:effect_of_schedule_strategies}.

\subsection{Sampling Based on Both Training Steps and Decoding Steps}
\label{sec:approach_both}
When comparing above two types of approaches, {\em i.e.,} $f(i)$ and $g(t)$, our approach $g(t)$ focus on simulating the distribution of real translation errors, and the vanilla $f(i)$ emphasizes the competence of the current model. 
Thus it is intuitive to verify whether $f(i)$ and $g(t)$ complement each other. 
How to combine them is the critical point.
At the training step $i$ and decoding step $t$, we define the probability of sampling golden tokens $ h(i, t)$ by the following joint distribution function:
\begin{itemize}
\item Product: $ h(i, t) = f(i) \cdot g(t)$
\item Arithmetic Mean: $ h(i, t) = \frac{f(i) + g(t)}{2}$
\item Composite\footnote{We also tried $f(i \cdot (1-g(t))$ in preliminary experiments, but it slightly underperformed the above $g(t \cdot (1-f(i)))$.}: $ h(i, t) = g(t \cdot (1-f(i)))$ 
\end{itemize}
One simple solution (`Product') is to directly multiply $f(i)$ and $g(t)$ . However, both $f(i)$ and $g(t)$ are less than or equal to 1, thus their product quickly shrinks to a tiny value close to 0.
Consequently, it exposes too few golden tokens and too many predicted tokens to the model (Figure \ref{fig:ss_on_both} (a)), which increases the difficulty for training.
`Arithmetic Mean' is another possible solution with a relatively gentle combination. However, it still inappropriately exposes too few golden tokens to the model at the beginning of training steps (Figure \ref{fig:ss_on_both} (b)).
Finally, we propose to apply function compositions on both $f(i)$ and $g(t)$ ({\em i.e.,} `Composite'). It guarantees enough golden tokens at the beginning of training steps, and gradually exposes more predicted tokens to the model with the increase of both $i$ and $t$ (Figure \ref{fig:ss_on_both} (c)).
We will analyze effects of different $h(i, t)$ in Section \ref{sec:sampling_on_both}.

\begin{table}[t!]
\begin{center}
\vspace{-5pt}
\scalebox{0.87}{
\begin{tabular}{l c c}
\hline \textbf{Dataset} & \textbf{Size (M)} & \textbf{Valid / Test set} \\
\hline
WMT14 EN-DE &  4.5 & newstest 2013 / 2014 \\
WMT14 EN-FR &  36 & newstest 2013 / 2014 \\
WMT19 ZH-EN &  20 & newstest 2018 / 2019 \\
\hline
CNN/DailyMail & 0.3 & standard data \\
Gigaword & 3.8 & standard data \\
\hline
\end{tabular}}
\end{center}
\vspace{-3pt}
\caption{Dataset statistics in our experiments.}
\label{table:dataset_statistics.}
\vspace{-5pt}
\end{table}

\begin{table*}[t!]
\begin{center}
\scalebox{0.91}{
\begin{tabular}{l l c c c c} 
\hline
\multirow{2}*{\textbf{Variable}} & \multirow{2}*{\textbf{Task}} & \multirow{2}*{\textbf{Maximum Value}}  &  \multicolumn{3}{c}{\textbf{Hyperparameter k}}  \\ \cline{4-6}
~ & ~ & ~ & \textbf{Linear } & \textbf{Exponential} & \textbf{Sigmoid}  \\
\hline
\multirow{2}*{Training Steps $f(i)$ (vanilla)} & Translation & 300,000  & -1/150,000 & 0.99999 & 20,000   \\
~ & Summarization & 100,000 & -1/50,000 & 0.9999 & 15,000   \\
\hline
\multirow{2}*{Decoding Steps $g(t)$ (ours)} & Translation & 128  & -1/64 & 0.99 & 20  \\
~ & Summarization & 512  & -1/256 & 0.999  & 50  \\
\hline
\end{tabular}}
\end{center}
\caption{Hyperparameters $k$ for different schedule strategies in our experiments.
}
\label{table:hyperparameters_for_ss}
\end{table*}

\section{Experiments}
We validate our proposals on two important sequence generation tasks, {\em i.e.,} machine translation and text summarization. 

\label{sec:experiments}
\subsection{Tasks and Datasets}
\paragraph{Machine Translation.}
We use the standard WMT 2014 English-German (EN-DE), WMT 2014 English-French (EN-FR), and WMT 2019 Chinese-English (ZH-EN) datasets. 
We respectively build a shared source-target vocabulary for EN-DE and EN-FR, and unshared vocabularies for ZH-EN.
We apply byte-pair encoding \cite{bpe_2016} with 32k merge operations for all datasets. 

\paragraph{Text Summarization.}
We use two popular summarization datasets: (a) the non-anonymized version of
the CNN/DailyMail dataset \cite{cnndw_dataset_2017}, and
(b) Gigaword corpus \cite{ggw_dataset_2015}.
We list dataset statistics for all datasets in Table \ref{table:dataset_statistics.}. 

\subsection{Implementation Details}
\label{subsec:imp_details}
\paragraph{Training Setup.}
For the translation task, we follow the default setup of the Transformer$_{base}$ and Transformer$_{big}$ models \cite{transformer_2017}, and provide detailed setups in Appendix A (Table \ref{table:params}).
All Transformer models are first trained by teacher forcing with 100k steps, and then trained with different training objects or scheduled sampling approaches for 300k steps. 
All experiments are conducted on 8 NVIDIA V100 GPUs,
where each is allocated with a batch size of approximately 4096 tokens.
For the text summarization task, we base on the ProphetNet \cite{prophetnet_2020} and follow its training setups. 
We set hyperparameters involved in various scheduled sampling strategies ({\em i.e.,} $f(i)$ and $g(t)$) according to the performance on validation sets of each tasks and list $k$ in Table \ref{table:hyperparameters_for_ss}.
For the linear decay, we set $\epsilon$  and $b$ to 0.2 and 1, respectively.
Please note that scheduled sampling is only used during training instead of the inference stage. 

\paragraph{Evaluation.}
For the machine translation task, we set the beam size to 4 and the length penalty to 0.6 during inference. We use \textit{multibleu.perl} to calculate cased sensitive BLEU scores for EN-DE and EN-FR, and use \textit{mteval-v13a.pl} script to calculate cased sensitive BLEU scores for ZH-EN.
We use the paired bootstrap resampling methods \cite{significance_test_2004} to compute the statistical significance of translation results. We report mean and standard-error variation of BLEU scores over three runs.
For the text summarization task, we respectively set the beam size to 4/5 and length penalty to 1.0/1.2 for Gigaword and CNN/DailyMail dataset following previous studies \cite{mass_2019,prophetnet_2020}. We report the F1
scores of ROUGE-1, ROUGE-2, and ROUGE-L for both datasets. 

\begin{table*}[t!]
\begin{center}
\scalebox{0.85}{
\begin{tabular}{l c c c c} 
\hline
\multirow{2}*{\textbf{Model}}  &  \multicolumn{3}{c}{\textbf{BLEU}}  \\ \cline{2-4}
& \textbf{EN-DE} & \textbf{ZH-EN} & \textbf{EN-FR}  \\
\hline
Transformer$_{base}$ \cite{transformer_2017} & 27.30 & ~~~~-- & 38.10   \\
Transformer$_{base}$ \cite{transformer_2017} $\dagger$ & 27.90 $\pm$ .02 & 24.97 $\pm$ .01 & 39.90 $\pm$ .02  \\
\ \ \  $+$ Mixer \cite{mixer_2015} $\dagger$ &  28.54 $\pm$ .02 & 25.28 $\pm$ .03 & 40.17 $\pm$ .01 \\
\ \ \  $+$ Minimal Risk Training  \cite{mrt_2016} $\dagger$   & 28.55 $\pm$ .01 & 25.33 $\pm$ .05 & 40.10 $\pm$ .02 \\
\ \ \  $+$ TeaForN \cite{teaforn_2020}  & 27.90 $\pm$ .03 & ~~~~-- & 40.84 $\pm$ .07  \\
\ \ \  $+$ TeaForN \cite{teaforn_2020} $\dagger$ & 28.60 $\pm$ .02 & 25.45 $\pm$ .02 & 40.34 $\pm$ .01 \\
\ \ \  $+$ Target denoising \cite{wechat_wmt_2020} $\dagger$  & 28.45 $\pm$ .02 & 25.78 $\pm$ .03 & 40.79 $\pm$ .02 \\
\ \ \  $+$ Sampling based on training steps \cite{ss_2015} $\dagger$  & 28.40 $\pm$ .01  & 25.43 $\pm$ .04  & 40.62 $\pm$ .03  \\
\ \ \  $+$ Sampling with sentence oracles  \cite{zhang_bridging_2019} & 28.65 & ~~~~--  & ~~~~-- \\
\ \ \  $+$ Sampling with sentence oracles  \cite{zhang_bridging_2019} $\dagger$  & 28.65 $\pm$ .03 & 25.50 $\pm$ .04 & 40.65 $\pm$ .02 \\
\ \ \  $+$ Sampling based on decoding steps (ours) $\dagger$ & ~~~~28.83 $\pm$ .05$**$ & ~~25.96 $\pm$ .07$*$ & ~~~~41.05 $\pm$ .04$**$ \\
\ \ \  $+$ Sampling based on training and decoding steps (ours) $\dagger$ & ~~~~\textbf{28.98 $\pm$ .03}$**$ & ~~~~\textbf{26.05 $\pm$ .04}$**$ & ~~~~\textbf{41.17 $\pm$ .03}$**$ \\
\hline
Transformer$_{big}$ \cite{transformer_2017} & 28.40 & ~~~~-- & 41.80 \\
Transformer$_{big}$ \cite{transformer_2017} $\dagger$ & 28.90 $\pm$ .03 & 25.22 $\pm$ .04 & 41.89 $\pm$ .03  \\
\ \ \  $+$ Mixer \cite{mixer_2015} $\dagger$& 29.27 $\pm$ .01  & 25.58 $\pm$ .02 & 42.37 $\pm$ .01  \\
\ \ \  $+$ Minimal Risk Training  \cite{mrt_2016} $\dagger$  & 29.35 $\pm$ .02 & 25.65 $\pm$ .01  & 42.46 $\pm$ .01  \\
\ \ \  $+$ TeaForN \cite{teaforn_2020}  & 29.30 $\pm$ .01 & ~~~~-- & 42.73 $\pm$ .01  \\
\ \ \  $+$ TeaForN \cite{teaforn_2020} $\dagger$  & 29.32 $\pm$ .01  & 25.48 $\pm$ .02  & 42.62  $\pm$ .01 \\
\ \ \  $+$ Error correction \cite{error_correction_2020} & 29.20 & ~~~~-- & ~~~~-- \\
 \ \ \  $+$ Target denoising \cite{wechat_wmt_2020} $\dagger$  & 29.68 $\pm$ .02 & 25.56 $\pm$ .03 & 42.62 $\pm$ .03 \\
\ \ \  $+$ Sampling based on training steps \cite{ss_2015} $\dagger$  & 29.62 $\pm$ .01  & 25.60 $\pm$ .02  & 42.55 $\pm$ .01  \\
\ \ \  $+$ Sampling with sentence oracles  \cite{zhang_bridging_2019} $\dagger$ & 29.57 $\pm$ .03  & 25.78 $\pm$ .02  & 42.65 $\pm$ .01  \\
\ \ \  $+$ Sampling based on decoding steps (ours) $\dagger$ & ~~29.85 $\pm$ .02$*$ & ~~~~\textbf{26.23 $\pm$ .01}$**$ & ~~~~42.87 $\pm$ .01$**$ \\
\ \ \  $+$ Sampling based on training and decoding steps (ours) $\dagger$ & ~~~~\textbf{30.16 $\pm$ .01}$**$  & ~~~~26.10 $\pm$ .01$**$ & ~~~~\textbf{43.13 $\pm$ .01}$**$ \\
\hline
\end{tabular}}
\end{center}
\caption{Translation performance of each dataset. `$\dagger$' is our implementations under unified settings. The original TeaForN \cite{teaforn_2020} reports SacreBLEU scores. For fair comparison, we re-implement it and report BLEU scores.
`$*$ / $**$': significantly better than vanilla `Sampling based on training steps' ($p < 0.05$ / $p < 0.01$).
}
\label{table:nmt_results}
\end{table*}

\subsection{Systems}

\paragraph{Mixer.}
A sequence-level training algorithm for text generations by combining both REINFORCE and cross-entropy \cite{mixer_2015}.

\paragraph{Minimal Risk Training.}
Minimal Risk Training (MRT) \cite{mrt_2016} introduces evaluation
metrics ({\em e.g.,} BLEU) as loss functions and aims to minimize expected loss on the training data.

\paragraph{Target denoising.}
\citet{wechat_wmt_2020} and \citet{wechat_wmt_2021} propose to add noisy perturbations into decoder inputs for a more robust translation model against prediction errors.

\paragraph{TeaForN.}
Teacher forcing with n-grams \cite{teaforn_2020} enables the standard teacher forcing with a broader view by a n-grams optimization.

\paragraph{Sampling based on training steps.}
For distinction, we name vanilla scheduled sampling as \textit{Sampling based on training steps}. We defaultly adopt the sigmoid decay following \citet{zhang_bridging_2019}.

\paragraph{Sampling with sentence oracles.}
\citeauthor{zhang_bridging_2019} (\citeyear{zhang_bridging_2019}) refine the sampling candidates of scheduled sampling with sentence oracles, {\em i.e.,} predictions from beam search. Note that its sampling strategy is based on training steps with the sigmoid decay.

\paragraph{Sampling based on decoding steps.}
 Sampling based on decoding steps with exponential decay.

\paragraph{Sampling based on training and decoding steps.}
Our sampling based on both training steps and decoding steps with the `Composite' method.

\subsection{Main Results}
\paragraph{Machine Translation.}
We list translation qualities on three WMT tasks in Table \ref{table:nmt_results}.
The sentence-level training based approaches ({\em e.g.,} Mixer) bring limited improvements due to the high variance of policy gradients and the credit assignment problem. On the contrary, sampling-based approaches show better translation qualities while preserving efficient training. TeaForN also yields competitive translation qualities due to its long-term optimization.
Among all existing methods, our `Sampling based on decoding steps' shows consistent improvements on various datasets.
Moreover, `Sampling based on training and decoding steps' combines the advantages of both existing methods and our proposals, and achieves better performance.
Specifically for the Transformers$_{base}$, it brings significant improvements by 1.08, 1.08, and 1.27 BLEU points on EN-DE, ZH-EN, and EN-FR, respectively.
Moreover, it significantly outperforms vanilla scheduled sampling by 0.58, 0.62, and 0.55 BLEU points on these tasks, respectively.
For the more powerful Transformers$_{big}$, we observe similar experimental conclusions as above. Specifically, `Sampling based on training and decoding steps' significantly outperforms the Transformers$_{big}$ by 1.26, 0.88 and 1.24 BLEU points on EN-DE, ZH-EN, and EN-FR, respectively. 

\begin{table*}[t!]
\begin{center}
\scalebox{0.92}{
\begin{tabular}{l c c c} 
\hline
\multirow{2}*{\textbf{Model}} &  \multicolumn{2}{c}{\textbf{RG-1 / RG-2 / RG-L}}  \\ \cline{2-3}
~  & \textbf{CNN/DailyMail} & \textbf{Gigaword}  \\
\hline
RoBERTSHARE$_{large}$ \cite{robertshare_2020}   & 40.31 / 18.91 / 37.62 & 38.62 / 19.78 / 35.94  \\
MASS \cite{mass_2019}                 & 42.12 / 19.50 / 39.01 & 38.73 / 19.71 / 35.96  \\
UniLM \cite{unilm_2019}               & 43.33 / 20.21 / 40.51 & 38.45 / 19.45 / 35.75  \\
PEGASUS$_{large}$ \cite{pegasus_2020}           & 44.17 / 21.47 / 41.11 & 39.12 / 19.86 / 36.24  \\
PEGASUS$_{large}$ + TeaForN \cite{teaforn_2020} & 44.20 / 21.70 / 41.32 & 39.16 / 20.16 / 36.54  \\
ERNIE-GEN$_{large}$ \cite{ernie_gen_2020}       & 44.31 / 21.35 / 41.60  & 39.46 / 20.34 / 36.74 \\
BART+R3F \cite{sota_summary_2020} (previous SOTA) & 44.38 / \textbf{21.53} / 41.17 & \textbf{40.45} / 20.69 / 36.56 \\
ProphetNet$_{large}$ \cite{prophetnet_2020} (primary baseline) $\dagger$  & 44.08 / 21.14 / 41/19 & 39.59 / 20.33 / 36.62  \\
 \ \ \  $+$ Target denoising \cite{wechat_wmt_2020} $\dagger$ & 43.98 / 21.09 / 41.08 & 39.68 / 20.18 / 36.78   \\
\ \ \  $+$ Sampling based on training steps \cite{ss_2015} $\dagger$ & 43.47 / 20.76 / 40.59 & 39.77 /  20.44 / 36.79  \\
\ \ \  $+$ Sampling based on decoding steps (ours) $\dagger$ & 44.20 / 21.33 / 41.41 & 40.11 / 20.39 / 37.15  \\
\ \ \  $+$ Sampling based on training and decoding steps (ours) $\dagger$ & \textbf{44.40} / 21.44 / \textbf{41.61} & 40.01 / \textbf{20.70} / \textbf{37.24}  \\

\hline
\end{tabular}}
\end{center}
\caption{F1 scores of ROUGE-1 / ROUGE-2 / ROUGE-L on test sets of both datasets. `RG' is short for `ROUGE'. `$\dagger$' is our implementations under a unified framework. 
Our approaches achieve consistently better performance.
}
\label{table:summary_results}
 \vspace{-3pt}
\end{table*}

\paragraph{Text Summarization.}
 In Table \ref{table:summary_results}, we list F1 scores of ROUGE-1 / ROUGE-2 / ROUGE-L on test sets of both text summarization datasets.
We take the powerful ProphetNet$_{large}$ as our primary baseline\footnote{The codes of previous SOTA \cite{sota_summary_2020} are not publicly available. Thus we base our approach on the second-best ProphetNet \cite{prophetnet_2020}.} and apply different sampling-based approaches.
For vanilla scheduled sampling  (second last row of Table \ref{table:summary_results}), we observe marginal improvements on Gigaword and even degenerations on CNN/DailyMail. 
We speculate that poor performance comes from their uniform sampling rate along with decoding steps, which violates the distribution of the real inference scene.
Namely, the model is overexposed to golden tokens and underexposed to predicted tokens at larger decoding steps.
Especially for CNN/DailyMail, its averaged target sequence length exceeds 64, and more than 90\% of sentences are longer than 50, which exacerbates the above issue in existing sampling-based approaches. We further analyze the effects of different sampling approaches on different sequence lengths in Section \ref{sec:effect_on_lengths}.
Nevertheless, our approaches are not affected by the above issue and show consistent improvements in all criteria of both datasets. 
Specifically, our approaches achieve consistently better performance than the baseline system on both datasets, and significantly improve the previous SOTA on ROUGE-L score of Gigaword to 37.24 (+0.5).
In conclusion, the strong performance on the text summarization task indicates that our approaches have a good generalization ability across different tasks.

\section{Analysis and Discussion}
In this section, we provide in-depth analyses on the necessity of our proposals and conducts experiments on the validation set of WMT14 EN-DE with the Transformer$_{base}$ model.

\subsection{Effects of Scheduled Strategies}
In this section, we focus on the effects of different scheduled strategies based on the decoding step $t$, and aim to answer the following two questions:

\begin{table}[t!]
\begin{center}
\scalebox{0.885}{
\begin{tabular}{c l c c}
\hline
\textbf{ID} & \textbf{Scheduled Strategies} & \textbf{ BLEU } & \textbf{ $\Delta$ } \\ 
\hline
\multirow{3}*{(a)} & No Sampling Baseline & 27.10 & ref. \\
~ & \ \ $+$ Always Sampling & 26.52 & -0.58 \\
~ & \ \ $+$ Uniform Sampling  & 27.48 & +0.38 \\
~ & \ \ $+$ Exponential Decay & \textbf{28.16} & \textbf{+1.06} \\
\hline
\multirow{4}*{(b)} & Uniform Sampling Baseline & 27.48 & ref. \\
~ & \ \ $+$ Linear Increase & 27.33 & -0.15 \\
~ & \ \ $+$ Exponential Increase & 27.25 & -0.23 \\
~ & \ \ $+$ Sigmoid Increase & 27.17 & -0.31  \\ 
\hline
\multirow{3}*{(c)}  & \ \ $+$ Linear Decay & 27.98 & +0.50 \\
~ & \ \ $+$ Exponential Decay & \textbf{28.16} & +\textbf{0.68} \\
~ & \ \ $+$ Sigmoid Decay & 28.05 & +0.57 \\
\hline
\end{tabular}}
\end{center}
 \vspace{-1pt}
\caption{BLEU scores (\%) on the validation set of WMT14 EN-DE for different schedule strategies $g(t)$. `ref.' indicates the reference baseline.}
\label{table:effects_of_strategies}
\vspace{-5pt}
\end{table}

\paragraph{(a) Is a Scheduled Strategy is Necessary?}
We take the Transformer without sampling as the baseline, then respectively apply `Always Sampling', `Uniform Sampling', and our `Exponential Decay'. Results are listed in the part (a) of Table \ref{table:effects_of_strategies}. We observe a noticeable drop when conducting `Always Sampling', as the model is entirely exposed to its predictions and fails to converge fully. 
As to `Uniform Sampling', it is essentially a simulation of the vanilla `Sampling based on Training Steps'. Although `Uniform Sampling' conducts an inappropriate sampling strategy, it still can simulate the data distribution of the inference scene to some extent and bring BLEU improvements modestly.
In contrast, our `Exponential Decay' conducts a sampling strategy following real translation errors. It significantly outperforms both `No Sampling' and `Uniform Sampling' by 1.64 and 0.68 BLEU scores. In short, we conclude that an appropriate scheduled strategy based on decoding steps is necessary.

\paragraph{(b) Why Decay Instead of Increase Strategies?}
Considering errors naturally accumulate along with decoding steps, both decay strategies and increase strategies can simulate error accumulations.
We respectively apply both kinds of sampling strategies upon the `Uniform Sampling' baseline model, and list results in the part(b) and part(c) of Table \ref{table:effects_of_strategies}. 
Surprisingly, all increase strategies consistently decrease performance by considerable margins. 
We conjecture that these increase strategies simulate an unreasonably high error rate at the beginning of decoding steps. Too many translation errors are propagated to subsequent decoding steps, which hinders the final performance.
On the contrary, all decay strategies bring consistent improvements with different degrees.
Moreover, we observe that the more a decay strategy approximates real error numbers (Figure \ref{fig:accumulated_errors}), the more performance improvements. In summary, we need to apply decay strategies instead of increase strategies based on decoding steps in the perspective of simulating real error accumulations.


\label{sec:effect_of_schedule_strategies}

\begin{table}[t!]
\begin{center}
\scalebox{0.82}{
\begin{tabular}{l c c}
\hline \textbf{Combination Methods} & \textbf{ BLEU } & \textbf{ $\Delta$ } \\ 
\hline
Sampling based on decoding steps & 28.16  & reference \\
\ \ \  $+$ Product & 27.65 & -0.51 \\
\ \ \  $+$ Arithmetic Mean & 28.06 & -0.10 \\
\ \ \  $+$ Composite & \textbf{28.37} & \textbf{+0.21} \\
\hline
\end{tabular}}
\end{center}
\caption{
BLEU scores (\%) on the WMT14 EN-DE validation set with different combination methods.
`Product' and `Arithmetic Mean' lead to performance degradation in different degrees. While `Composite' can further improve the strong baseline to a certain extent.
}
 \vspace{-2pt}
\label{table:ss_on_both}
\end{table}






\subsection{Effects of Different $h(i, t)$ Strategies}
\label{sec:sampling_on_both}
We take our strong `Sampling based on decoding steps' as the baseline and then apply different combination methods $h(i, t)$. 
As shown in Table \ref{table:ss_on_both}, the performance drop of `Product' and `Arithmetic Mean' confirms our speculation in Section \ref{sec:approach_both}.
Namely, the model is overexposed to its predictions at the beginning of training steps and decoding steps, thus fails to converge well.
In contrast, `Composite' brings certain improvements over the strong baseline model. Since it stabilizes the model training and successfully combines the advantages of both dimensions of training steps and decoding steps. 
In summary, a well-designed strategy is necessary when combining both $f(i)$ and $g(t)$, and we provide an effective alternative ({\em i.e.,} `Composite').

\begin{figure}[t!]
\begin{center}
     \scalebox{0.48}{
      \includegraphics[width=1\textwidth]{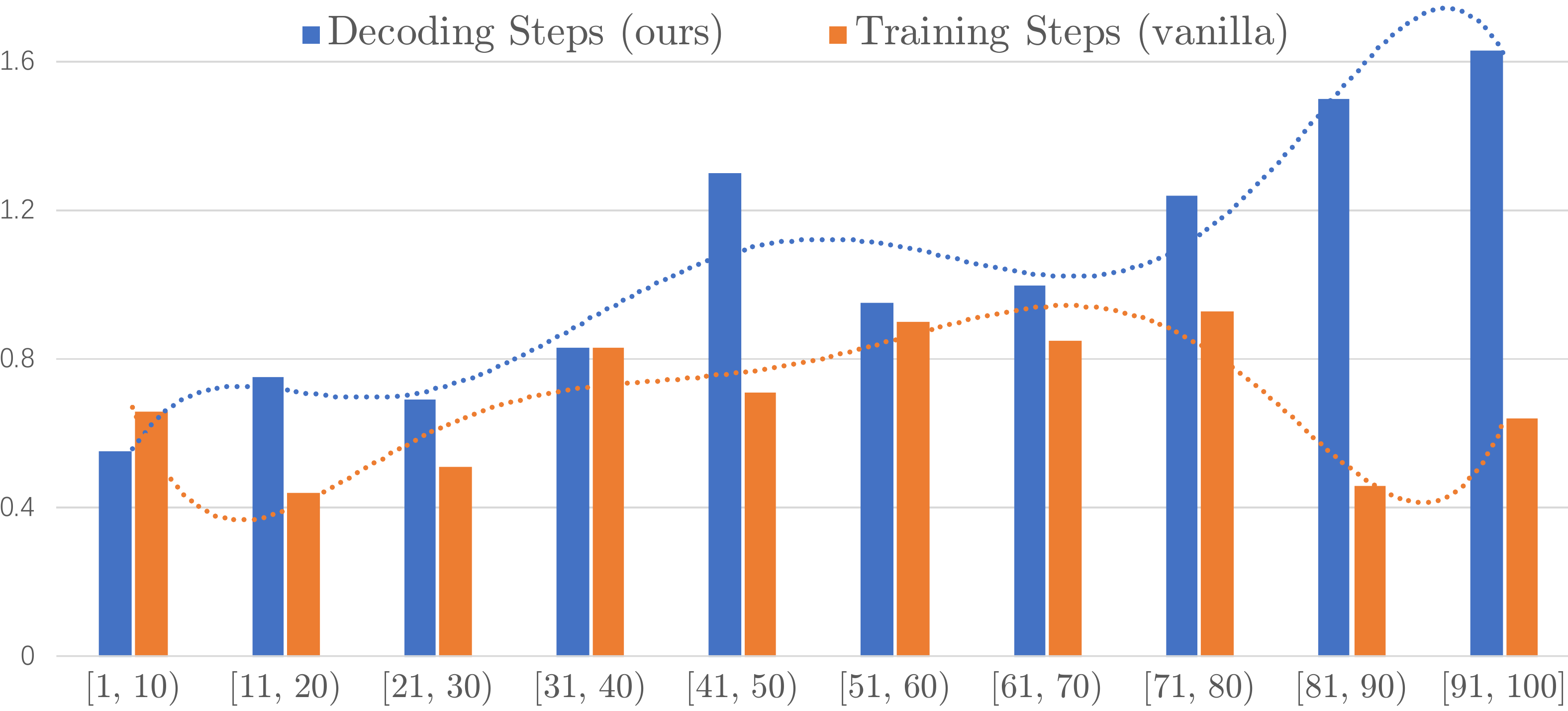}
      } 
      \caption{Absolute BLEU scores (\%) gains over the Transformer baseline on different sequence lengths, where dashed lines are polynomial trendlines.
      } 
      \label{fig:gain}  
 \vspace{-10pt}     
 \end{center} 
\end{figure}

\subsection{Effects on Different Sequence Lengths}
\label{sec:effect_on_lengths}
 According to our early findings, the exposure bias problem gets worse as the sentence length grows. Thus it is intuitive to verify whether our approaches improve translations of long sentences. Since the size of WMT14 EN-De validation set (3k) is too small to cover scenarios with various sentence lengths, we randomly select training data with different sequence lengths. Specifically, we divide WMT14 EN-DE training data into ten bins according to the source side’s sentence length. The maximal length is 100, and the interval size is 10. Then we randomly select 1000 sentence pairs from each bin and calculate BLEU scores for different approaches. Specifically, we take the Transformer as the baseline, and draw absolute BLEU gains of scheduled sampling on training steps and decoding steps. As shown in Figure~\ref{fig:gain}, BLEU gains of the vanilla scheduled sampling are relatively uniform over different sentence lengths. In contrast, BLEU gains of our scheduled sampling on decoding steps gradually increase with sentence lengths. Moreover, our approach consistently outperforms the vanilla one at most sentence length intervals. Specifically, we observe more than 1.0 BLEU improvements when sentence lengths in $[80; 100]$.

\section{Conclusion}
  In this paper, we propose scheduled sampling methods based on decoding steps from the perspective of simulating real translation error rates, and provide in-depth analyses on the necessity of our proposals.
  We also confirm that our proposals are complementary with existing studies (based on training steps).
  Experiments on three large-scale WMT translation tasks and two text summarization tasks confirm the effectiveness of our approaches. 
   In the future, we will investigate low resource settings which may suffer from a more serious error accumulation problem. In addition, more autoregressive-based tasks would be explored as future work.

  
 
  
  
  

\section*{Acknowledgements}
The research work described in this paper has been supported by the National Key R\&D Program of China (2020AAA0108001) and the National Nature Science Foundation of China (No. 61976015, 61976016, 61876198 and 61370130).  
 Jinan Xu is the corresponding author of the paper.

\bibliography{anthology,custom}
\bibliographystyle{acl_natbib}

\newpage
\clearpage

\appendix
\section{Training Details}
\label{sec:training_details}
We list detailed parameters for training Transformer models in Table \ref{table:params}.

\begin{table}[t!]
\begin{center}
\scalebox{0.8}{
\begin{tabular}{l c c}
\hline \textbf{Parameter} & \textbf{Transformer$_{base}$} & \textbf{Transformer$_{big}$} \\ 
\hline
batch size & 4096 & 4096 \\
number of GPUs & 8 & 8 \\ 
hidden size  & 512  & 1024 \\
filter size  & 2048  & 4096 \\
number of heads & 8 & 16 \\
number of encoders & 6 & 6 \\
number of decoders & 6 & 6 \\
dropout & 0.1 & 0.3 \\
label smoothing & 0.1 & 0.1 \\
pre-training steps & 100,000 & 200,000 \\
fine-tuning steps & 300,000 & 300,000 \\
warmup steps & 4,000 & 8,000 \\
learning rate & 1.0 & 1.0 \\
optimizer & Adam & Adam \\
Adam beta1 & 0.9 & 0.9 \\
Adam beta2 & 0.98 & 0.98 \\
layer normalization & post-norm & post-norm \\
position encoding & absolute & absolute \\
share embeddings & True & True \\
share softmax weights & True & True \\ 
\hline
\end{tabular}}
\end{center}
\caption{Detailed parameters for Transformer$_{base}$ and Transformer$_{big}$ on all WMT datasets. Note that the `share embedding' is set to `False' on WMT19 ZH-EN.}
\label{table:params}
\end{table}

\section{Real Error Rates as Sampling Priors}
In the above contents of this paper, we aim to better simulate the inference scene under the guidance of real error rates. We can not help wondering the effect of directly taking the above error rates as sampling priors. Disappointingly, it fails to outperform our exponential decay strategy within a gap of 0.1 BLEU scores. 
We conjecture the metric we used to measure translation errors at each decoding step may not be good enough.
Considering the optimal metric is currently unknown and unavailable, our unigram matching can yet be regarded as a simple and effective alternative. 
It succeeds in reflecting the trend of real error rates and brings significant improvements by simulating the error distribution estimated by unigram matching. 
We believe a better metric would bring further improvements and leave this exploration for future work.

\end{document}